\documentclass[letterpaper, 10 pt, conference]{ieeeconf}  

\IEEEoverridecommandlockouts                              

\usepackage{multirow}
\usepackage{booktabs}
\usepackage{graphicx}
\usepackage{url}
\usepackage{colortbl}

\usepackage{xcolor}
\usepackage{hyperref}
\usepackage{cite}
\usepackage{amsfonts}
\usepackage[utf8]{inputenc}
\usepackage{amsmath}
\usepackage{algorithm}
\newcommand{\LeftComment}[1]{\Statex \hspace{\algorithmicindent}/* #1 */}
\newcommand{\greyrule}{\arrayrulecolor{black!30}\midrule\arrayrulecolor{black}}
\usepackage{algpseudocode}

\title{\LARGE \bf
Enhancing Indoor Mobility with Connected Sensor Nodes: A Real-Time, Delay-Aware Cooperative Perception Approach
}

\author{Minghao Ning$^{1*}$, Yaodong Cui$^{1*}$, Yufeng Yang$^{1}$, Shucheng Huang$^{1}$, Zhenan Liu$^{1}$ \\
Ahmad Reza Alghooneh$^{1}$, Ehsan Hashemi$^{2}$ and Amir Khajepour$^{1}$
\thanks{$*$ Equal contribution.}
\thanks{$^{1}$Minghao Ning, Yaodong Cui, Shucheng Huang, Zhenan Liu, Ahmad Reza Alghooneh and Amir Khajepour are with the Mechanical and Mechatronics Eng. Department, University of Waterloo, 200 University Ave W, Waterloo, ON N2L3G1, Canada. e-mail:\{minghao.ning, yaodong.cui, f248yang, s95huang, z634liu, aralghoo, a.khajepour\}@uwaterloo.ca).}
\thanks{$^{2}$Ehsan Hashemi is with the Mechanical Engineering Department, University of
Alberta, Alberta, T6G1H9, Canada (e-mail:ehashemi@ualberta.ca)}%
}

\begin{document}

\maketitle
\thispagestyle{empty}
\pagestyle{empty}

\begin{abstract}
This paper presents a novel real-time, delay-aware cooperative perception system designed for intelligent mobility platforms operating in dynamic indoor environments. The system contains a network of multi-modal sensor nodes and a central node that collectively provide perception services to mobility platforms. The proposed Hierarchical Clustering Considering the Scanning Pattern and Ground Contacting Feature based Lidar Camera Fusion improve intra-node perception for crowded environment. The system also features delay-aware global perception to synchronize and aggregate data across nodes. To validate our approach, we introduced the Indoor Pedestrian Tracking dataset, compiled from data captured by two indoor sensor nodes. Our experiments, compared to baselines, demonstrate significant improvements in detection accuracy and robustness against delays. The dataset is available in the repository\footnote{\url{https://github.com/NingMingHao/MVSLab-IndoorCooperativePerception}}.

\end{abstract}



\section{Introduction}
In recent years, intelligent indoor autonomy technology is gaining recognition and attention among healthcare professionals and researchers. Studies have shown that indoor transportation is the most urgent need from healthcare staff in hospitals and long-term care \cite{hebesberger2015staff}. This rising demand is largely driven by workforce shortages and the high incidence of chronic injuries among healthcare staff, which often caused by transporting heavy materials. However, large scale commercial deployment of intelligent robotics platforms are still limited. Most existing indoor robots are designed to operate independently, relying on their built-in sensors to navigate and perform tasks. This restricts their effectiveness in the congested, dynamic, and unpredictable spaces of healthcare facilities. This paper presents a cooperative perception system consisting of a network of multiple sensor nodes, and a central node, to provide perception results/services to robotic mobility platforms. This system aimed to improve the operational safety and environmental awareness of intelligent robotic platforms, including autonomous hospital beds and delivery robots.

\begin{figure}[t]
    \centering
    \includegraphics[width=0.45\textwidth]{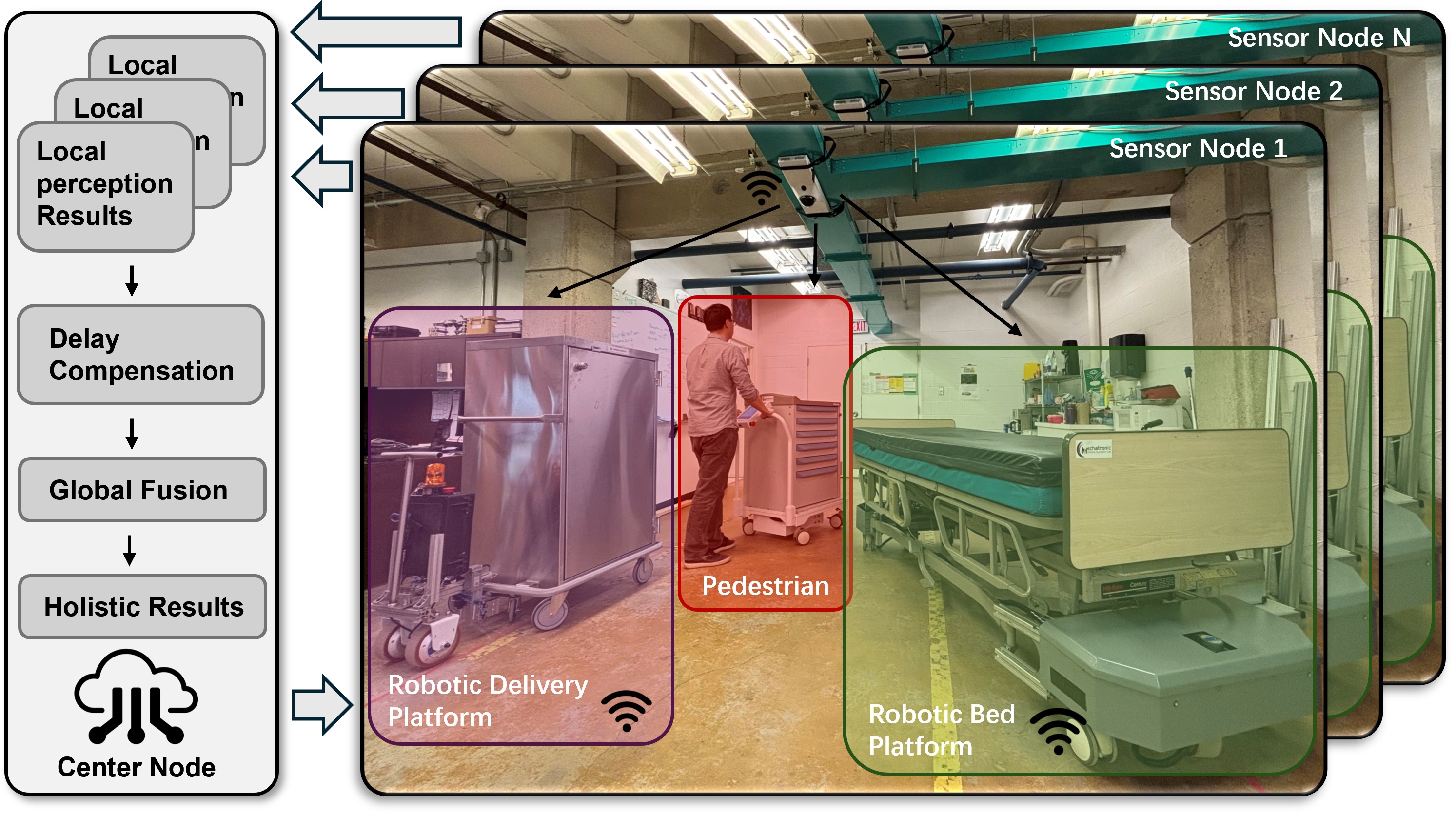}
    \vspace*{-0.1cm}
    \caption{Overview of the proposed cooperative perception system}
    \label{fig:overview}
\vspace*{-0.4cm}
\end{figure}

There are several challenges associated with developing a cooperative perception system in densely populated indoor environments, such as hospitals. One primary challenge for local perception is the fast and accurate fusion of perception data from multiple sensor nodes. This task is complicated by the dynamic behavior of people within a confined space, which involves close interactions between individuals. For instance, people travel in small groups, or crossing paths at close quarters. These situations pose significant difficulties in maintaining consistent tracking identities across different nodes and merging perception data effectively. The physical layout of indoor environments presents another significant challenge for local perception. Architectural features and decorative elements, such as corners, pillars, and mirrors, present significant challenges in achieving continuous and accurate coverage across the entire area. These environmental factors can obstruct the sensor field of view and distort the sensor signal, leading to gaps in coverage or inaccuracies in perception.
\begin{figure*}[!ht]
    \centering
    \includegraphics[width=0.8\textwidth]{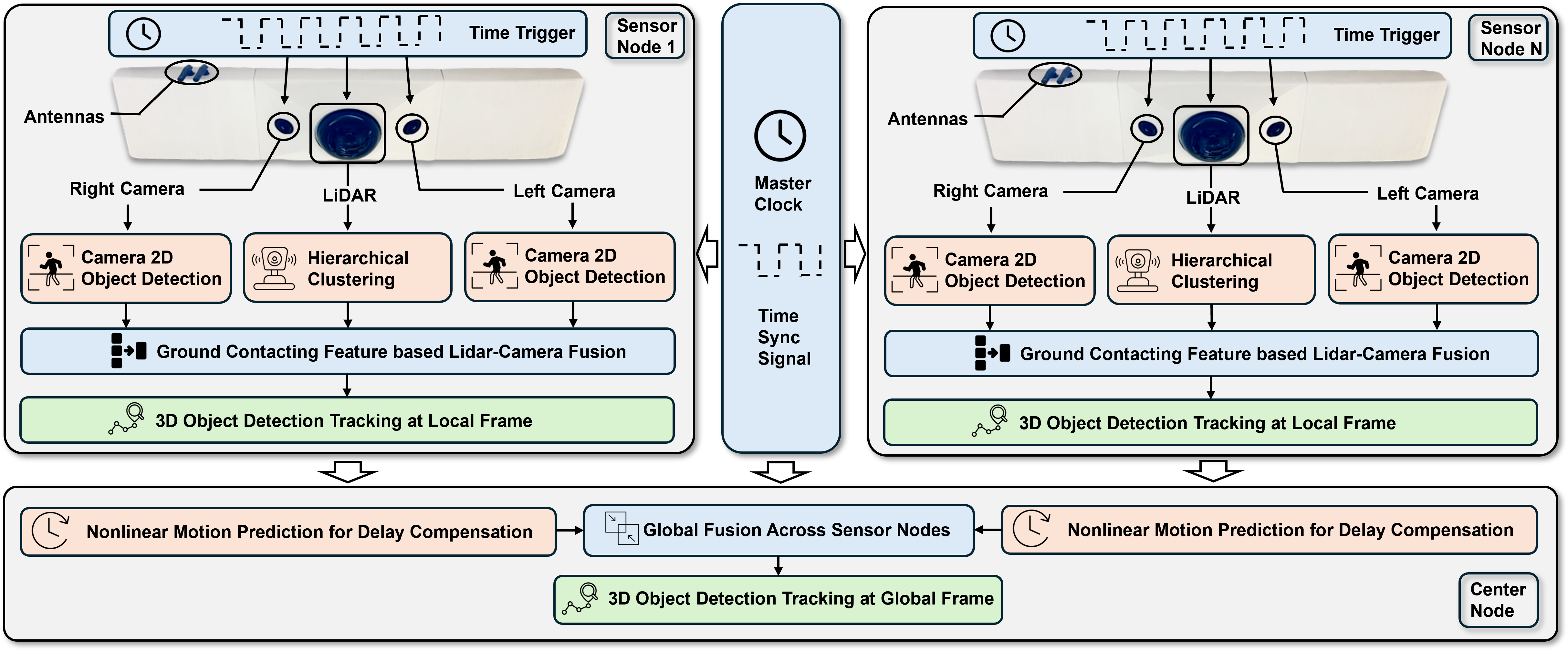}
    \vspace*{-0.1cm}
    \caption{The proposed delay-aware cooperative perception framework.}
    \label{fig:framework}
    \vspace*{-0.4cm}
\end{figure*}

The processing and communication delays poses a major challenge for global/cross-sensor perception in highly dynamic indoor environments. These cross-node delays can lead to the receipt of outdated or inaccurate representations of the dynamic environment at the center node. This impairs the center node's ability to generate a cohesive and current understanding of the environment. 

To address these challenges, this paper proposes a delay-aware cooperative perception system designed for dynamic indoor environment. An overview of the proposed system is illustrated in Fig. \ref{fig:overview}. Our contribution can be summarized as follows:
\begin{itemize}

\item An adaptive clustering method coupled with ground-contact point-based LiDAR-camera fusion, enhancing the accuracy and reliability of local perception.

\item A delay-aware global perception framework that accounts for messaging delays and latency, ensuring timely and cohesive environmental understanding.

\item The creation of a multimodal cooperative indoor perception dataset specifically designed for dynamic and crowded healthcare environments. This provids a valuable resource for further research and development in this field.
\end{itemize}

The rest of the paper is organized as follows, in section II, the related methods and dataset are reviewed, in section III, the overview of our method is presented, in section IV, the experiments and discussion are presented, and finally in section V the impact of our work is concluded.

\section{Related Work}

\subsection{Indoor Perception}
Existing indoor infrastructure-based perception system often relies on basic sensors and cameras, which either lack high-level semantic understanding or precise measurement of object positions. In \cite{wu2021cooperative}, four Pyroelectric Infrared (PIR) sensors are combined as a sensor node and mounted on the ceiling to detect object trajectory. In \cite{5504205} Radio frequency identification (RFID) is used to track objects embedded with RFID tags. Although these methods provide basic tracking functionalities, their perception range and accuracy are very limited. In \cite{zhou2012understanding,heya2018image,haque2018visionbased} infrastructure-based cameras are used to detect and track pedestrians. However, such standalone pure vision-based systems are sensitive to lighting variations and occlusions and cannot accurately localize objects in 3D space. Alternatively, Brvsvcic et al. leveraged a combination of infrastructure-based RGB-D cameras, LiDAR, and marker-based motion tracking systems \cite{brvsvcic2013person}. However, the cost of such setup makes them impractical for large-scale deployment. A more recent study used a motion capture system capable of producing ground truth data at a 100 Hz rate \cite{rudenko2020thor}. Despite its high accuracy, it is limited to areas where motion capture technology is available. These challenges highlight the need for more robust and cost-effective perception systems capable of operating reliably under the complex conditions typical in indoor settings.

\subsection{Indoor Cooperative Perception Dataset}
\begin{table}[htb]
\scriptsize
\centering
\caption{Indoor dataset summary (Inf: Infrastructure, Auto: Automated, GT: Ground truth) }
\label{tab:indoor_dataset}
\begin{tabular}{@{}llll@{}}
\toprule
\textbf{Dataset}                   & \textbf{Mounting}            & \textbf{Sensors}    &  \textbf{Annotation} \\ \midrule
KTH \cite{dondrup2015real} & Robot & RGB-D, 2D Lidar  & Auto\\  
L-CAS \cite{yan2017online} & Robot & 3D LiDAR  & Manual\\ 
MuSoHu \cite{nguyen2023toward} & Wearable  & RGB/Stereo camera, 3D LiDAR & N/A\\ 
Central station \cite{zhou2012understanding}  & Inf& Camera & Auto\\ 
ATC \cite{brvsvcic2013person} & Inf & RGB-D, Lidar, Motion tracking  & Auto\\ 
Thor \cite{rudenko2020thor}  & Inf & Motion capture &  GT\\
\textbf{MVSL}(Our) & Inf & Camera, 3D Lidar &  GT\\
\bottomrule
\end{tabular}
\end{table}
As shown in Table~\ref{tab:indoor_dataset}, existing indoor datasets typically rely on RGB and depth cameras, LiDARs, and motion capture/tracking systems to obtain the position of each object. Despite their utility, these datasets fail to fully capture the scope of indoor environments and dynamics due to the inherent limitations of the technologies employed. For instance, cameras (RGB-D) and LiDAR sensors installed on mobile robots or wearable devices \cite{dondrup2015real, yan2017online, nguyen2023toward} are limited by their range, FOV, and issues like object truncation and occlusions.

\section{Methodology}
As shown in Fig. \ref{fig:framework}, the proposed delay-aware cooperative system comprises two main components: local perception for sensor nodes and delay-aware global perception on a center node. Each sensor node is equipped with dual cameras, a LiDAR sensor, 5G/wireless communication capabilities, and a Jetson Orin NX for edge computing. These nodes process multi-modal sensory data locally to produce tracked object lists. By integrating edge computing capabilities, we aim to reduce the overall system latency. The center node aggregates and combines the structured perception results of the sensor nodes to generate a holistic view of the dynamic indoor environment. This configuration allows for real-time detection and tracking of dynamic elements across multiple nodes in complex indoor settings.

\subsection{Local Perception}
The local perception can be summarized into: cross-node sensor synchronization; camera based 2D bounding box detection; ROI points filtering; hierarchical clustering considering the scanning pattern;  ground contacting feature based Lidar camera fusion; and class-aware object tracking. 

\subsubsection{Cross-sensor Sensor Synchronization}
\begin{figure}[t]
    \centering
    \includegraphics[width=0.49\textwidth]{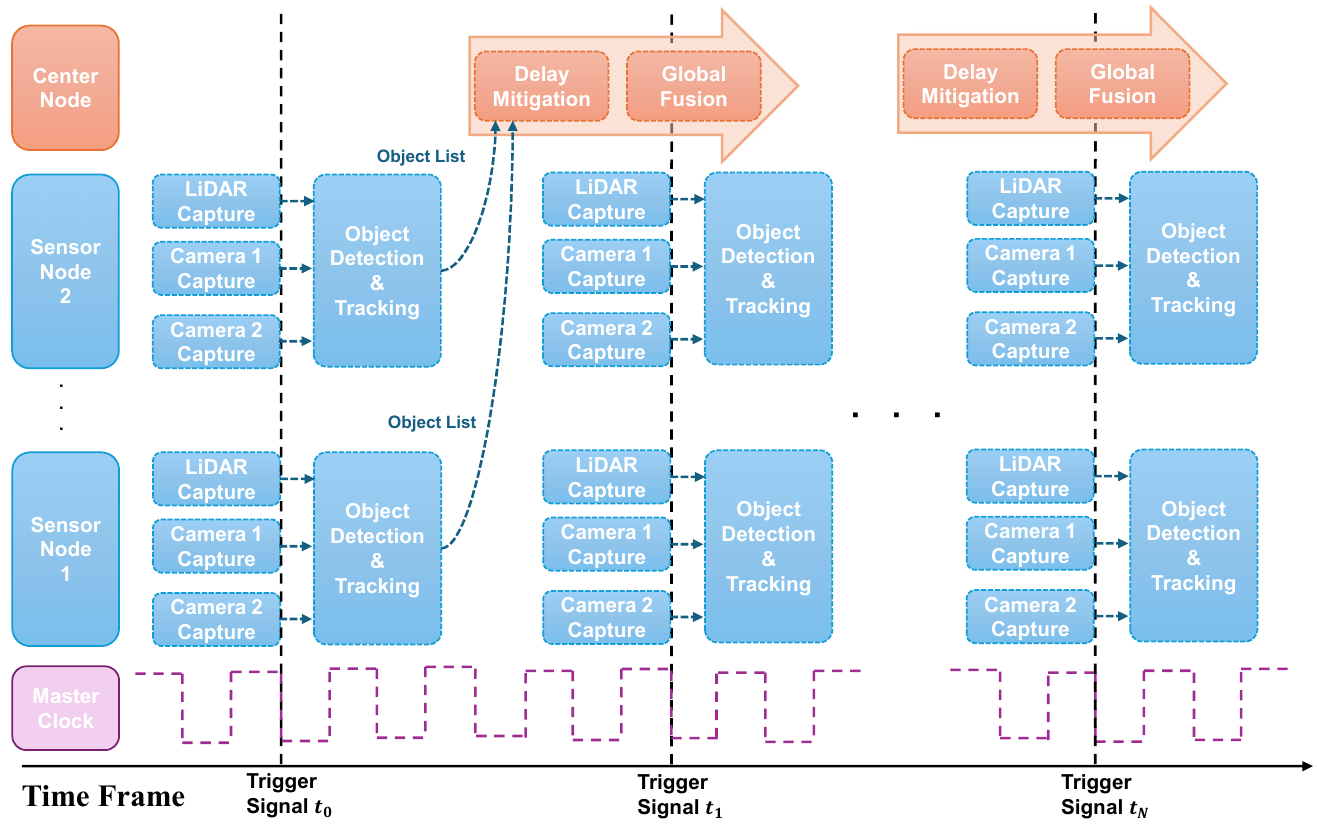}
    \vspace*{-0.2cm}
    \caption{\textbf{The time synchronization process.} \textbf{Master clock.} ensure uniform trigger signals for simultaneous data capture. \textbf{Sensor Node} soft triggers ensures temporal alignment of multi-modal data. \textbf{Center Node} aggregation and processing of synchronized data from all nodes.}
    \label{fig:time-sync}
    \vspace*{-0.4cm}
\end{figure}

To improve the accuracy of global fusion, the proposed framework employs cross-sensor soft synchronization mechanism to reduce delay in the captured and processed data. As shown in Fig.\ref{fig:time-sync}, each sensor node coordinates LiDAR scans with camera shutter operations through the use of soft trigger signals. This trigger signal ensures that the data captured from both modalities are temporally aligned. The generation of these soft trigger signals is based on a synchronized clock system. Each node's clock is synchronized to ensure the uniformity of trigger signals across all sensor nodes. This alignment allows for simultaneous data capture between nodes. This synchronization significantly reduces discrepancies during the global fusion process and improves the accuracy of the global perception's output. 

\subsubsection{Camera-based 2D Detection}
For camera-based 2D object detection, we employ a custom YOLOv8\cite{yolov8_ultralytics} model trained on our dataset. Standard YOLO models trained on COCO(Common Objects in Context) dataset\cite{lin2015microsoft} doesn't generalize well on the proposed infrastructure view, and it can not detect the ground contacting features (like foot for person). So a customized dataset including person, foot, and robot bed labels are created to retrain the YOLOv8 and evaluate its performance.
\subsubsection{ROI points filtering}
As the sensor nodes are fixedly installed, a static binary grid is created as the region of interest to filter out unnecessary points, such as points on the wall or ground. 

\subsubsection{Hierarchical Clustering Considering the Scanning Pattern}
\begin{figure}[t]
    \centering
    \includegraphics[width=0.485\textwidth]{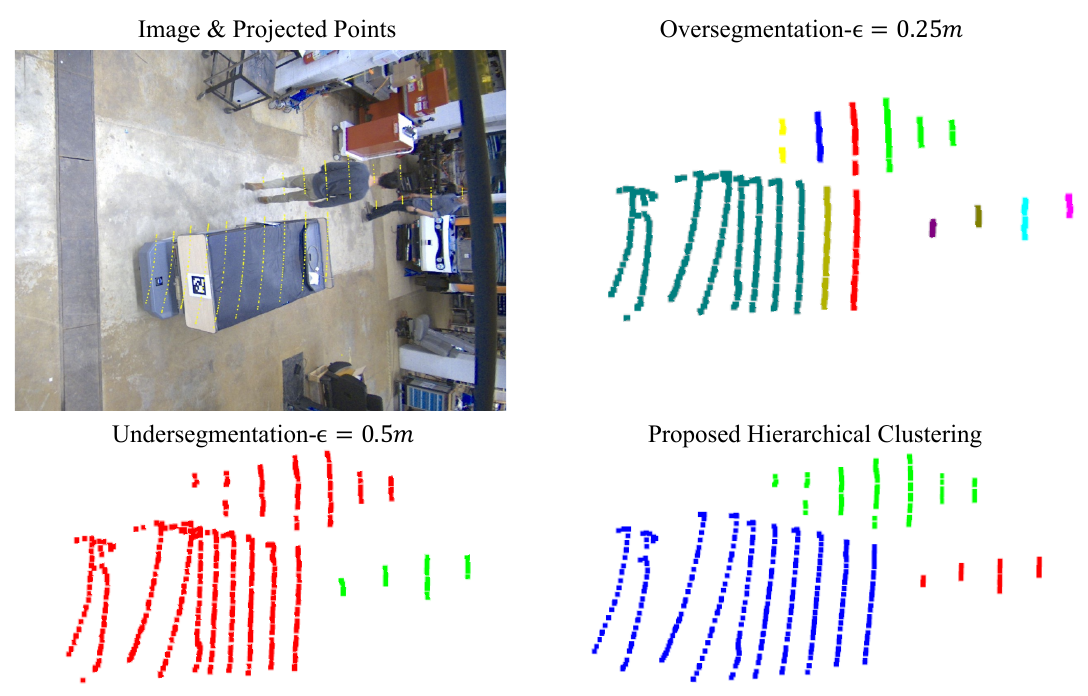}
    \vspace*{-0.4cm}
    \caption{\textbf{Clustering Example.} \textbf{Image and the projected points.} \textbf{Over-Segmentation-$\epsilon=0.25m$.} \textbf{Under-Segmentation-$\epsilon=0.5m$.}  \textbf{Proposed Hierarchical Clustering.} \textbf{Different clusters are shown in different colors.}} 
    \label{fig:cluster-example}
    \vspace*{0.1cm}
\end{figure}

Common clustering methods like DBSCAN \cite{DBSCAN} assume the points are spatially uniformly distributed, where the Euclidean distance between points of the same cluster should scale equally along different axes of the Cartesian Coordinates. However, this assumption fails for the wildly used mechanical rotating Lidar, where the resolution along the horizontal direction is much finer than that along the vertical direction. Thus, careful clustering parameter tuning for the clustering distance threshold $\epsilon$ is required to have a better trade-off between the under-segmentation and the over-segmentation issues. As shown in Fig.\ref{fig:cluster-example}, large $\epsilon$ tends to occur under-segmentation, and small $\epsilon$ leads to over-segmentation. However, no proper $\epsilon$ exists in this case that can properly cluster the two persons and the robot bed, as the distance between the upper right corner of the bed and its nearby person is smaller than the distance between the points from the two nearby scanning lines at the right side of the bed.

To address the above issue, an efficient hierarchical clustering method considering the scanning pattern is proposed. First, points from different scanning lines are clustered separately based on the adaptive euclidean distance $\epsilon (s)$, 
\begin{equation}
    \epsilon (s)= N_{\min} \Delta \varphi s
\end{equation}
where $s$ is the distance from the point to the Lidar, $N_{\min}$ is the minimum number of points required to form a core region in DBSCAN, and $\Delta \varphi$ is the horizontal resolution of the Lidar.

\begin{figure}[t]
    \centering
    \includegraphics[width=0.485\textwidth]{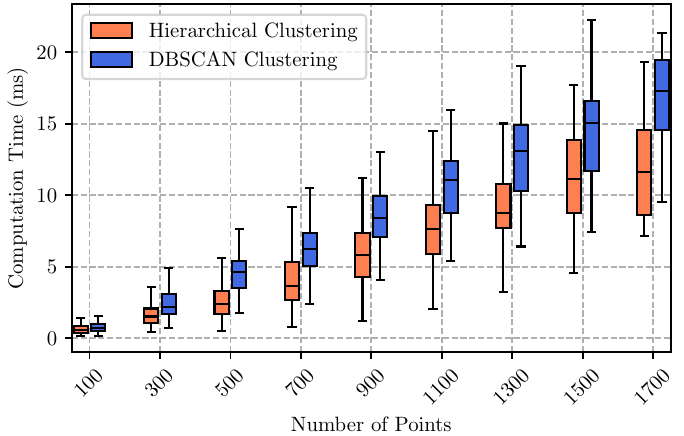}
    \vspace*{-0.6cm}
    \caption{\textbf{Computation Time Comparison.}} 
    \label{fig:clustering-time}
    \vspace*{-0.3cm}
\end{figure}

Then, a custom distance metric considering the scanning pattern is proposed to group the segments of each scanning lines from the first step. Each segment contains the points and other features like the ring index of the scanning line, the centroid calculated as the mean of the cluster points, the azimuth angle $\varphi$ range denoting the start and the end scanning angle for this segment. The custom distance for any two segments is calculated based on Algorithm \ref{alg:clustering}. The segments whose distance is less than threshold $\epsilon_{custom}$ will be grouped into the same cluster. 

It is worth noting that this hierarchical clustering is faster than the DBSCAN, as distance calculation across different scanning lines has reduced from point-to-point to segment-to-segment. This improvement greatly reduce the computational time when the number of points increases as shown in Fig. \ref{fig:clustering-time}.

\begin{algorithm}[t]
\caption{Efficient Hierarchical Clustering Considering Scanning Patterns}
\label{alg:clustering}
\begin{algorithmic}[1]
\Function{DistanceMetric}{$segment\_a$, $segment\_b$}
    \State \textbf{Input:} $segment\_a$, $segment\_b$ - segments with properties (ring index, mean point, $\varphi$ range)
    \State \textbf{Output:} $distance$ - customized distance metric
    \LeftComment{Preliminary checks to speed up computation:}
    \If {large ring index or mean point distance}
        \State \Return $INF$
        \Comment{Segments too far apart in ring index or spatially}
    \EndIf
    \LeftComment{Calculate custom distance metric:}
    \State Compute spatial distance $d$ between mean points and normalize it to get $d_{norm}=d/(\min({s_a,s_b})\Delta \theta)$, where $\Delta \theta$ is the vertical resolution
    \State Compute $\varphi$ angle intersection $\varphi_\cap$ and normalize it to get $\varphi_{\cap norm}=1-\varphi_\cap / (\min(||\varphi_a||, ||\varphi_b||))$
    \State Compute distance $d_{custom}=d_{norm}+\varphi_{\cap norm}$
    \State
    \Return Custom distance metric $d_{custom}$
\EndFunction
\end{algorithmic}
\end{algorithm}

\subsubsection{Ground Contacting Feature based Lidar Camera Fusion}
The camera-based 2D detection results are fused with the pointcloud clustering results to assign semantic labels to the clusters. The fusion of 2D bounding box and point cloud clusters is challenging when objects are crowded, which create occlusion on the image view. For instance, when a group of people travel closely together. To solve the fusion problem in a cluttered scene, a ground contacting feature-based Lidar camera fusion method is proposed. Specifically, the camera projection matrix as shown in Eqn.\ref{eq:camera-projection} is used to estimate the actual position of the object in the world coordinate based on the detected 2D bounding box. 
\begin{equation}
    s \begin{bmatrix} x_p \\ y_p \\ 1 \end{bmatrix} = K [R \ t] \begin{bmatrix} x_w \\ y_w \\ z_w \\ 1\end{bmatrix} = H_{3\times4} \begin{bmatrix} x_w \\ y_w \\ z_w \\ 1\end{bmatrix}
    \label{eq:camera-projection}
\end{equation}
where $s$ is a scale factor, $x_p$, $y_p$ are the pixel coordinates, $K$ is the camera intrinsic matrix, $R$ is the rotation matrix, $t$ is the translation vector, $x_w$, $y_w$, $z_w$ are the world coordinates.

The ground contacting feature bounding box (like foot) will be first associated with its parent bounding box (like person) based on the bounding box overlap ratio and cosine distance with the $z$-axis vanishing point $v_z$, i.e., the pixel coordinate where all lines parallel to the $z$-axis in the world coordinates intersect. The overlap ratio is computed as the area of intersection between two boxes, divided by the minimum area of the two boxes. The cosine distance is the cosine of the angle between the vectors from the vanishing point to the centroids of two boxes. The vanishing point $v_z$ is calculated based on the camera matrix $H$
\begin{equation}
    v_z = (h_{13}/h_{33}, h_{23}/h_{33})
    \label{eq:vz}
\end{equation}

Then, the actual position, i.e. $x_w$ and $y_w$, of the object is derived based on Eqn.\ref{eq:camera-projection} given its pixel coordinates $x_p$, $y_p$ and its height $z_w$:
\begin{equation}
\begin{aligned}
a_{11} = h_{31}x_p - &h_{11}, \ a_{12} = h_{32}x_p - h_{12}\\
a_{21} = h_{31}y_p - &h_{21}, \ a_{22} = h_{32}y_p - h_{22}\\
b_{1} = (h_{13} - &h_{33}x_p)z_w + h_{14} - h_{34}x_p\\
b_{2} = (h_{23} - &h_{33}y_p)z_w + h_{24} - h_{34}y_p\\
x_w = &\frac{b_1 a_{22} - b_2 a_{12}}{a_{11}a_{22} - a_{12}a_{21}} \\
y_w = &\frac{b_2 a_{11} - b_1 a_{21}}{a_{11}a_{22} - a_{12}a_{21}}
\end{aligned}
\label{eq:actual-position-estimation}
\end{equation}

Finally, the 2D boxes are associated with the clusters using Hungarian algorithm to minize the overall association costs. The association cost for each pair of 2D box and cluster is the sum of the overlap ratio of camera box and the projected cluster box, and the Euclidean distance between the 2D box based estimated position and its cluster centroid.


\subsection{Delay-aware Global Perception}
To address the inherent challenges associated with the processing and communication of high volumes of data in real-time, our framework incorporates a delay-aware fusion algorithm within the center node. This algorithm utilizes precise timestamps from the detected object lists received from the sensor network. It then compares these with the current time to assess the delay encountered during data transmission from the sensor nodes to the central node. The central node then predicts the current positions of the detected objects based on type-based motion models. 

For pedestrian class objects, we use a non-linear motion model Eq. \ref{eq:motion} that considers both the speed and direction of movement, allowing the model to anticipate changes in a person’s trajectory. This prediction is important for improving the accuracy of cross-node fusion, especially in complex scenarios involving multiple dynamic objects in close proximity. After delay compensation, the center node combines these adjusted object lists by applying a weighted fusion strategy. This process ensures an accurate and up-to-date representation of the environment despite delays in data transmission.


\begin{equation}
\begin{aligned}
x_{k+1} &= x_k + v_{x_k} \cos(\text{yaw}_k) \Delta t, \\
y_{k+1} &= y_k + v_{x_k} \sin(\text{yaw}_k) \Delta t, \\
\text{yaw}_{k+1} &= \text{yaw}_k + \omega_{z_k} \Delta t, \\
v_{x_{k+1}} &= v_{x_k}, \\
\omega_{z_{k+1}} &= \omega_{z_k}.
\end{aligned}
\label{eq:motion}
\end{equation}

\section{Experiments}
\subsection{Dataset Overview and Metrics}
To assess the performance of the proposed algorithms, the Indoor Pedestrian Tracking dataset is created using data gathered from two indoor sensor nodes. It comprises 3,248 frames, featuring up to nine pedestrians and one hospital bed, with a total number of 22,857 objects labeled as tracked objects using CVAT \cite{cvat}. On average, there are 7.04 objects per frame in this dataset.

In detail, it consists of three distinct scenarios: 1) a challenging case with nine pedestrians, testing the algorithm's ability to handle high pedestrian traffic; 2) a scenario with four pedestrians, allowing for detailed analysis of tracking precision; and 3) a unique setting that includes a hospital bed and three pedestrians, focusing on the interaction between an autonomous hospital bed and humans in medical or assisted-living environments.

For the data labeling, LiDAR point clouds are initially filtered based on height and ROI, then cropped to remove ground points. The resultant point clouds are projected into a bird's-eye view for data labeling. Finally, the position and orientation of objects are labeled as bounding boxes on the bird's-eye view images.



For evaluation, precision, recall, and average distance error (Avg. DE) are adopted to assess the accuracy of object detection.

\subsection{Local Perception Evaluation}
\begin{table}[h]
\scriptsize 
\setlength{\tabcolsep}{4pt}
\centering
\caption{Local Perception Evaluation Results. DBSCAN1 has a lower $N_{\min}$, and DBSCAN2 has a higher $N_{\min}$.}
\begin{tabular}{c|c|cccc}
\hline
\textbf{Scenario} & \textbf{Node} & \textbf{Method} & \textbf{Precision} & \textbf{Recall} & \textbf{Avg. DE} \\
\hline
\multirow{6}{*}{9 people} & \multirow{3}{*}{Node1} & DBSCAN1 & 0.7679 & 0.9529 & 0.08539 \\
                          &                        & DBSCAN2 & 0.9652 & 0.9361 & 0.0846 \\
                          &                        & Our     & \textbf{0.9696} & \textbf{0.966}  & \textbf{0.0716} \\
\cline{2-6}
                          & \multirow{3}{*}{Node2} & DBSCAN1 & 0.6347 & 0.8827 & 0.0783 \\
                          &                        & DBSCAN2 & \textbf{0.9891} & 0.7355 & 0.0721 \\
                          &                        & Our     & 0.9649 & \textbf{0.9773} & \textbf{0.0692} \\
\hline
\multirow{6}{*}{4 people} & \multirow{3}{*}{Node1} & DBSCAN1 & 0.7061 & \textbf{0.9479} & 0.0885 \\
                          &                        & DBSCAN2 & 0.9463 & 0.9138 & 0.0864 \\
                          &                        & Our     & \textbf{0.9663} & 0.9461 & \textbf{0.0815} \\
\cline{2-6}
                          & \multirow{3}{*}{Node2} & DBSCAN1 & 0.5467 & 0.8378 & 0.0816 \\
                          &                        & DBSCAN2 & \textbf{0.981}  & 0.6425 & \textbf{0.0769} \\
                          &                        & Our     & 0.9607 & \textbf{0.9761} & 0.0779 \\
\hline
\multirow{6}{*}{3 people, 1 bed} & \multirow{3}{*}{Node1} & DBSCAN1 & 0.3231 & 0.9681 & 0.1087 \\
                                 &                        & DBSCAN2 & 0.9415 & 0.9441 & 0.0896 \\
                                 &                        & Our     & \textbf{0.9516} & \textbf{0.9944} & \textbf{0.0842} \\
\cline{2-6}
                                 & \multirow{3}{*}{Node2} & DBSCAN1 & 0.5909 & 0.8948 & 0.1369 \\
                                 &                        & DBSCAN2 & 0.9697 & 0.7283 & 0.0998 \\
                                 &                        & Our     & \textbf{0.9745} & \textbf{0.9881} & \textbf{0.0748} \\
\hline
\end{tabular}
\label{tab:eva_result_local}
\end{table}






The results of the local perception evaluation, as depicted in Table \ref{tab:eva_result_local}, show the comparative performance of two DBSCAN configurations against our proposed method. DBSCAN1 utilizes a parameter setting of $\epsilon=0.3m$ and $N_{\min}=4$, contrasting with DBSCAN2's configuration of $\epsilon=0.3m$ and $N_{\min}=8$.

Within the context of the \textbf{9 people} scenario, our approach significantly outperforms the competing methodologies, achieving a precision of 0.9696 and a recall of 0.966 for Node1, coupled with a precision of 0.9649 and a recall of 0.9773 for Node2. These results suggest superior performance in scenarios characterized by crowded conditions and potential occlusions. Despite DBSCAN2 achieving marginally greater precision in Node2, it suffers a considerable drop in recall, highlighting a limitation in detecting all relevant objects within a crowded environment.

In the \textbf{4 people} scenario, our method sustains high levels of both precision and recall, underscoring its efficacy. In contrast, normal DBSCAN experiences a compromise between precision and recall, which suggests its limitation to balance object detection with false positive mitigation effectively.

The \textbf{3 people, 1 bed} scenario introduces substantial challenges to standard DBSCAN configurations, particularly affecting DBSCAN1, where the difference in point densities leads to a notable drop in precision. This can be attributed to the 
oversegmentation issues, where the bed is erroneously clustered into multiple groups, resulting in an increased false positive rate and consequently, reduced precision. Conversely, our method demonstrates consistent high precision and recall across this scenario, underscoring its resilience in environments with variable point densities.

The average distance error is another critical factor in evaluating the performance of these methods, with our method exhibiting lower Avg. DE values across the majority of scenarios and nodes. This metric further demonstrates the spatial accuracy of our method in object localization tasks.


\subsection{Delay Mitigation}
\begin{figure}[t]
    \includegraphics[width=0.45\textwidth]{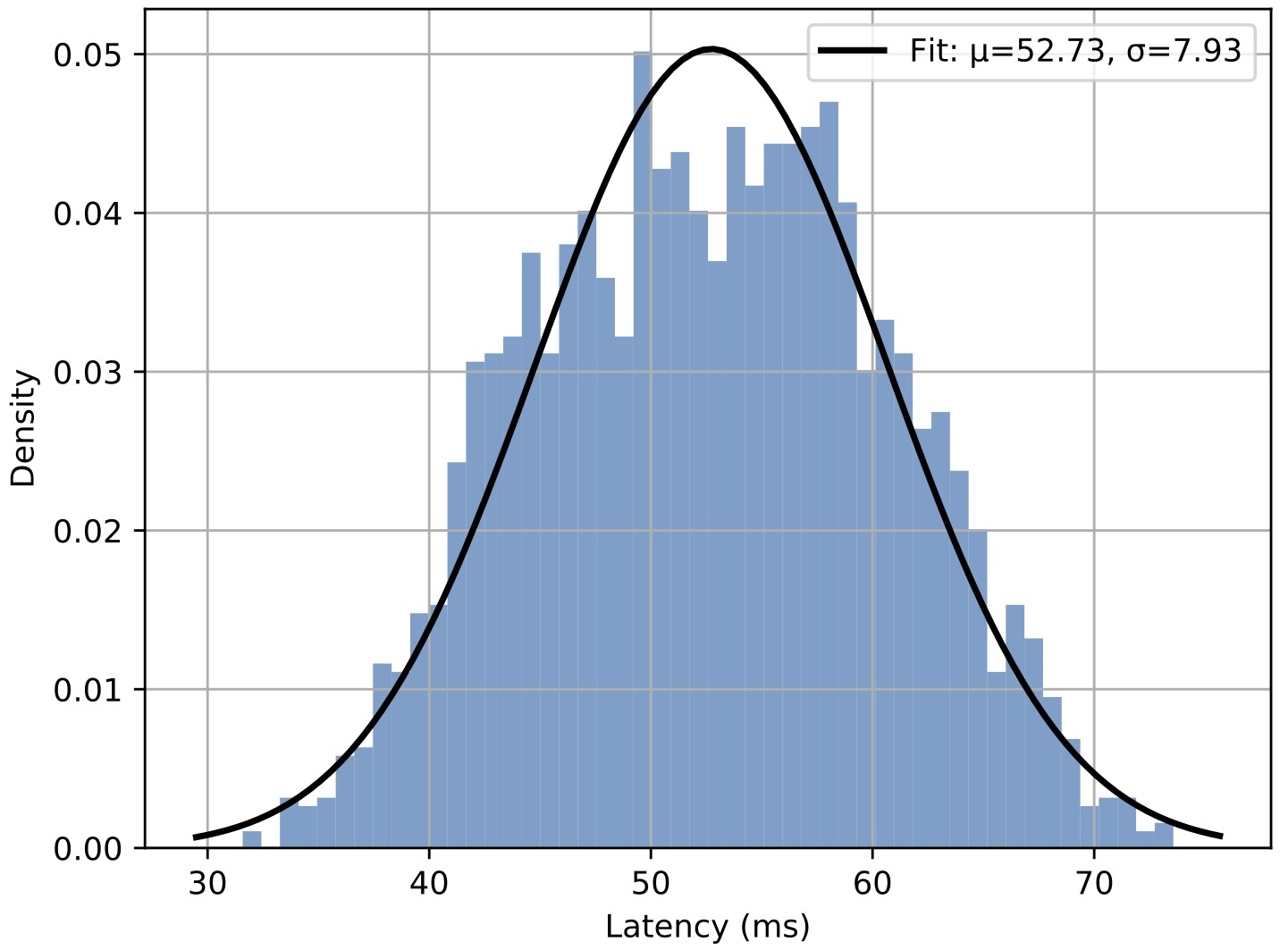}
    \vspace*{-0.3cm}
    \caption{\textbf{Recorded 5G Latency Distribution and its distribution fitting.}} 
    \label{fig:5g-latency}
    \vspace*{-0.2cm}
\end{figure}

The latency distribution for the communication between the sensor node and the center node over 5G is depicted in Fig. \ref{fig:5g-latency}. This distribution can be approximated by a Gaussian model, with a mean latency of 52.7 ms and a standard deviation of 7.9 ms. In the experiment, we first simulate this latency distribution with a mean of 50 ms and a standard deviation of 8 ms. To further explore the delay effects on system performance, we then mean latency to 100 ms and 150 ms, while keeping the standard deviation unchanged.


We compare our proposed delay mitigation method with a baseline method under three simulated latency configurations, the results are summarized in Table \ref{tab:eva_result_delay}. The delay-aware method consistently outperformed the baseline in terms of precision and average distance error across all scenarios and delay settings. This trend becomes more evident as the delay increased, with the delay-aware system maintained an averaged 18\% precision improvement over the baseline. In scenarios with fewer dynamic elements, the improvements were still noticeable, although the differences in recall were less consistent. For example, in the \textbf{3 people} scenario with a 100 ms delay, although the recall decreased slightly from 0.7338 to 0.7002, the precision saw a significant increase from 0.8053 to 0.8701.


\begin{table}[h]
\centering
\scriptsize
\caption{Delay Mitigation evaluation results. }
\begin{tabular}{@{}cccccc@{}}
\toprule
\textbf{Scenario}  & \textbf{Delay} & \textbf{Method} & \textbf{Precision} & \textbf{Recall} & \textbf{Avg. DE} \\ \midrule
\multirow{6}{*}{9 people} & \multirow{2}{*}{50ms}  & Baseline      & 0.8641 & \textbf{0.9072} & 0.2162 \\ 
                                                 &  & Delay-Aware  & \textbf{0.8943} & 0.8771 & \textbf{0.1267} \\ 
                           & \multirow{2}{*}{100ms}  & Baseline    & 0.7495 & 0.7196 & 0.2629 \\ 
                                                 &  & Delay-Aware  & \textbf{0.8799} & \textbf{0.7262} & \textbf{0.1330} \\ 
                           & \multirow{2}{*}{150ms}  & Baseline   & 0.7006 & 0.7160 & 0.2734 \\ 
                                                 &  & Delay-Aware  & \textbf{0.8626} & \textbf{0.7663} & \textbf{0.1492} \\ \greyrule 
\multirow{6}{*}{4 people} & \multirow{2}{*}{50ms}  & Baseline      & 0.8220 & \textbf{0.7891} & 0.2260 \\ 
                                                 &  & Delay-Aware  & \textbf{0.8567} & 0.7463 & \textbf{0.1315} \\ 
                           & \multirow{2}{*}{100ms}  & Baseline    & 0.6938 & 0.5867 & 0.2735 \\ 
                                                 &  & Delay-Aware  & \textbf{0.8289} & \textbf{0.6201} & \textbf{0.1494} \\ 
                           & \multirow{2}{*}{150ms}  & Baseline   & 0.6308 & 0.5889 & 0.2986 \\ 
                                                 &  & Delay-Aware  & \textbf{0.8175} & \textbf{0.6461} & \textbf{0.1472} \\   \greyrule      
\multirow{6}{*}{3 people, 1 bed} & \multirow{2}{*}{50ms}  & Baseline & 0.8663 & \textbf{0.9353} & 0.1997 \\ 
                                                 &  & Delay-Aware  & \textbf{0.8930} & 0.8715 & \textbf{0.1218} \\ 
                           & \multirow{2}{*}{100ms}  & Baseline    & 0.8053 & \textbf{0.7338} & 0.2306 \\ 
                                                 &  & Delay-Aware  & \textbf{0.8701} & 0.7002 & \textbf{0.1368} \\ 
                           & \multirow{2}{*}{150ms}  & Baseline   & 0.7502 & \textbf{0.7709} & 0.2510 \\ 
                                                 &  & Delay-Aware  & \textbf{0.8648} & 0.7596 & \textbf{0.1346} \\ \bottomrule
\end{tabular}
\label{tab:eva_result_delay}
\end{table}

\subsubsection*{Discussion on Delay Mitigation}
The improved performance of the delay-aware method can be attributed to its capability to compensate for network-induced delays, thereby improving the accuracy of object fusion and synchronization across sensors. This is particularly important in densely populated environments where precise localization is necessary for safe and effective robot navigation. The reduction in average distance error also indicates the system's ability to align data temporally.

Variations in recall of the proposed method are caused by duplicate objects after fusion. When pedestrians change direction unexpectedly in regions where sensor nodes overlap, motion prediction can result in incorrect fusion outcomes. This trade-off between detection coverage (recall) and detection accuracy (precision) is a common challenge in real-time perception systems and warrants further investigation to optimize both aspects.

\section{Conclusion}
This paper presented a cooperative perception system designed for intelligent mobility platforms in dynamic indoor settings, focusing on healthcare facilities. Our system integrates a network of multi-modal sensor nodes with a central node to address the challenges of crowded and unpredictable environments. We introduced novel algorithm designs, such as hierarchical clustering considering scanning patterns, ground contacting feature-based LiDAR camera fusion and delay-aware perception. The proposed approach significantly improves detection accuracy and operational safety, critical in crowded indoor settings. Experimental results from the Indoor Pedestrian Tracking dataset demonstrate our system's advantages over traditional baselines in terms of detection precision and delay robustness.

Future research will aim to extend this proposed framework to the transportation setting, such as traffic intersection or a specific section of road. 

\section*{Acknowledgement}
The authors would like to acknowledge the financial support of the Natural Sciences and Engineering Research Council of Canada (NSERC), MITACS and financial and technical support of Rogers Communications Inc.

\bibliographystyle{IEEEtran}
\bibliography{main}


\end{document}